\documentclass{article}
\usepackage[T1]{fontenc}
\usepackage{amssymb}
\usepackage{amsmath}
\usepackage{graphicx}
\usepackage{subcaption}
\usepackage{booktabs} 
\usepackage{enumitem}
\usepackage{tabularx}
\usepackage{float}
\usepackage[final]{corl_2026} 
\hypersetup{pdftitle={Eye, Robot: Learning Expressive Gaze Dynamics via a VLM-Guided Transformer-VAE},
  pdfauthor={Yuhang Hu, Junlin Lyu, Yuan Fang, Xiaoying Shen, Lechen Zhang, Jiong Lin, Judah Goldfeder, Steven Tian, Hod Lipson}}

\title{Eye, Robot: Learning Expressive Gaze Dynamics via a VLM-Guided Transformer-VAE}

\author{
  {\large\bfseries Yuhang Hu\textsuperscript{1,*}, Junlin Lyu\textsuperscript{1,*,\textdagger}, Yuan Fang\textsuperscript{1},}\\
  {\large\bfseries Xiaoying Shen\textsuperscript{1}, Lechen Zhang\textsuperscript{1}, Jiong Lin\textsuperscript{1},}\\
  {\large\bfseries Judah Goldfeder\textsuperscript{1}, Steven Tian\textsuperscript{1}, Hod Lipson\textsuperscript{1,\textdagger}}\\[3pt]
  {\normalsize\textsuperscript{1}Creative Machines Lab, Department of Mechanical Engineering, Columbia University}\\
  {\footnotesize\textsuperscript{*}\,Equal contribution.}
}

\begin{document}
\maketitle
\begingroup
\renewcommand{\thefootnote}{\fnsymbol{footnote}}
\footnotetext[2]{Corresponding authors: Junlin Lyu (\texttt{jl6963@columbia.edu}); Hod Lipson (\texttt{hl2891@columbia.edu}).}
\endgroup


\begin{abstract}

Gaze coordinates attention in human interaction through coupled eye, head, and eyelid motion. A humanoid robot must choose where to look while generating a suitable movement toward that target; fixed tracking rules alone do not model the temporal structure of human gaze. We separate these tasks with an asynchronous vision-language model that selects a scene referent, a reactive perception module that maintains fixation, and a Realistic Gaze Transformer (RGT) that generates target-conditioned motion. RGT is a Transformer-VAE trained on markerless human recordings retargeted to the robot's 11-DoF joint space: self-supervised pretraining encodes recent full-head history, and autoregressive fine-tuning exposes the model to its own predictions during rollout. At inference, a four-dimensional binocular-eye trajectory conditions coordinated neck and eye motion; an auxiliary blink predictor triggers eyelid templates during playback. On a two-second benchmark in which all controllers receive the same ground-truth target, RGT-Base obtains the lowest reported overall seven-joint MSE, and physical demonstrations show the integrated targeting and control pipeline. The evidence establishes robot-space kinematic fidelity and real-time RGT inference; perceived naturalness and open-vocabulary target-selection accuracy remain questions for controlled evaluation.

\end{abstract}



\section{Introduction}

Gaze is a fundamental channel of nonverbal communication, regulating attention, conversational turn-taking, intimacy, and emotional signaling \cite{frischen2007gaze,stephenson2021gaze,kompatsiari2017importance,pfeiffer2013gaze,argyle1965eye,kendon1967some}. Appropriate eye contact has further been shown to increase perceived trustworthiness \cite{10.1145/2522628.2522630}. In Human-Robot Interaction (HRI), embodiment allows robots to emulate these social cues through physical motion \cite{fong2003survey,breazeal2003toward,bartneck2004design,dautenhahn2007socially,hu2024human,chen2021smile}, which makes faithfully reproducing human gaze both a substantial challenge and a promising path toward more natural interaction \cite{admoni2017social}.

\begin{figure}[h!]
    \begin{center}
    \includegraphics[width=0.99\textwidth]{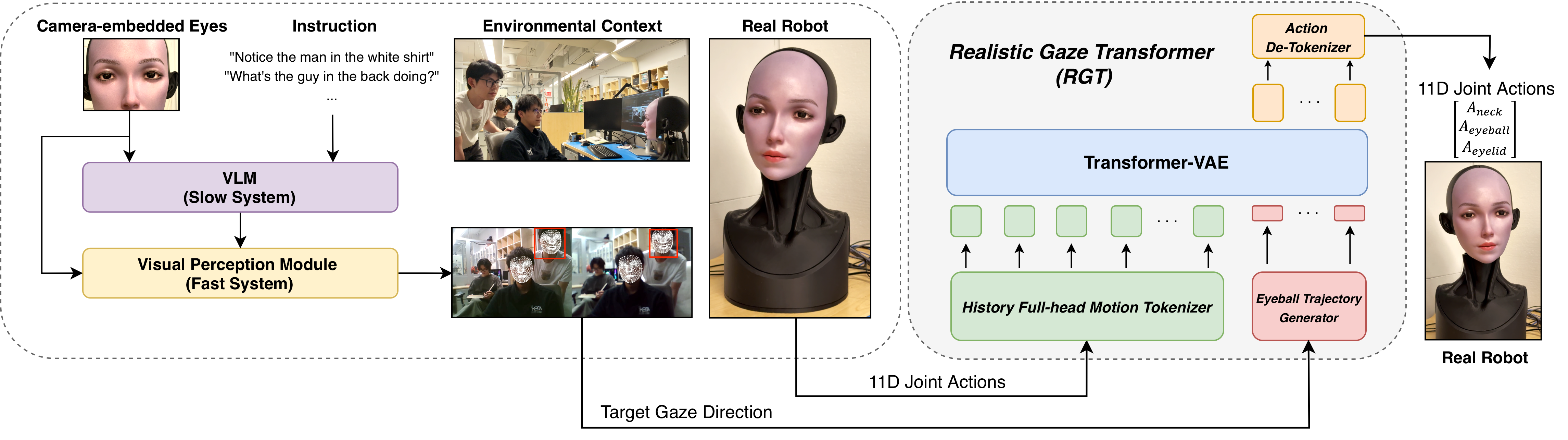}
    \end{center}
    \vspace{-10pt}
    \caption{\textbf{Overview of the Realistic Gaze Transformer (RGT) pipeline.} 
    The left panel illustrates our hierarchical perception system operating on a 
    real humanoid robot. Egocentric video from the camera-embedded eyes, together 
    with high-level natural language instructions, is processed by a slow-system VLM that reasons over 
    the environmental context and designates salient targets. A fast-system 
    Visual Perception Module then tracks the selected faces in real time to 
    produce a continuous target gaze direction. The right panel depicts the RGT 
    architecture: a History Full-head Motion Tokenizer encodes recent facial 
    dynamics while an Eyeball Trajectory Generator encodes the target gaze, and a 
    Transformer-VAE fuses both streams into a latent representation from which it 
    predicts future motion. The Action De-Tokenizer maps the near-future tokens 
    to 11-DoF joint commands on the physical robot. The quantitative benchmark
    evaluates motion generation with ground-truth eye targets; target selection
    is illustrated by the physical demonstration.}
    \vspace{-10pt}
    \label{fig:pipeline}
\end{figure}

Scripted eye and head controllers provide a useful starting point, but fixed rules have limited flexibility across targets and motion styles~\cite{mutlu2012conversational,admoni2014deliberate,pan2020realistic}. Gaze also influences first impressions in social interaction~\cite{paetzel2020persistence,zlotowski2015anthropomorphism}, making the timing of robot movements worth studying alongside endpoint tracking. We therefore ask whether a prior learned from human motion can improve joint-space reproduction under target-conditioned, closed-loop robot control (Fig.~\ref{fig:pipeline}). The framework separates a deliberative decision about the attended referent from the frame-rate generation of motor commands. A VLM selects a referent asynchronously, a reactive perception module maintains its fixation between VLM calls, and the Realistic Gaze Transformer (RGT) predicts coordinated neck, binocular-eye, and eyelid commands from recent motion and a planned eye trajectory.

The action model is pretrained on retargeted human recordings and then fine-tuned autoregressively, so training includes histories containing its own predictions rather than only recorded frames. Its latent bottleneck summarizes the motion history while the target trajectory enters the decoder after that bottleneck. The 11-DoF output combines seven neck/eye joints from the learned motion path with four eyelid joints and a separate blink-template playback path. This differs from generating body motion for an avatar: commands must remain in the robot's joint coordinates, track a binocular target, and roll out at the control rate. The hierarchy permits a slow semantic decision and faster reactive control without requiring the VLM to issue each joint command. It also lets us examine motion generation separately from errors in choosing or tracking a referent.

Our contributions are (i) a VLM-guided, reactive targeting interface for a humanoid face, (ii) a target-conditioned Transformer-VAE with a latent bottleneck and autoregressive fine-tuning for coordinated robot gaze, and (iii) a markerless data and evaluation pipeline that tests closed-loop fidelity, stochastic diversity, participant adaptation, and runtime. The quantitative benchmark supplies every controller with the same ground-truth eye target and scores seven neck/eye channels; physical demonstrations show the integrated targeting pipeline. Code and simulation materials are linked at \url{https://anonymous.4open.science/r/eye_robot_pub-CF64/}. 

\section{Related Work}
\label{sec:citations}

\textbf{Learning on Humanoid Facial Robots.} Early facial robots relied on
hand-tuned, FACS-based presets~\cite{breazeal2003toward,faraj2021facially}, which
scale poorly and feel mechanical. Recent work shifts to learning-based control
across facial affect: self-supervised mimicry from motor
babbling~\cite{chen2021smile}, expression generation guided by recognition
networks~\cite{rawal2022exgennet} or human preferences~\cite{yang2025hapi},
anticipatory smile coexpression~\cite{hu2024human}, and multilingual lip
synchronization~\cite{hu2026lipmotions}. Gaze, a primary channel for attention and
social engagement, has received less attention within this learning paradigm. Our
setting specifically models synchronized robot-coordinate neck, binocular-eye,
and eyelid trajectories with a target supplied after the latent bottleneck.

\textbf{Robot Gaze Control.} Early systems scripted eye and head movements by hand to mimic human attention patterns \cite{mutlu2012conversational, admoni2014deliberate}. Recognizing the limits of this approach, later work turned to bio-inspired and model-based strategies---such as smooth pursuit, saccades, and joint attention---to coordinate eye and head motion for more lifelike behavior \cite{roncone2016cartesian,mishra2022knowing,moon2014meet,pan2020realistic}, and Saran et al.~\cite{saran2018human} used a monocular deep-learning framework for real-time gaze following. Despite this progress, realistic gaze remains an open challenge, which motivates control that learns directly from human motion.

\begin{figure}[h!]
\includegraphics[width=0.99\textwidth]{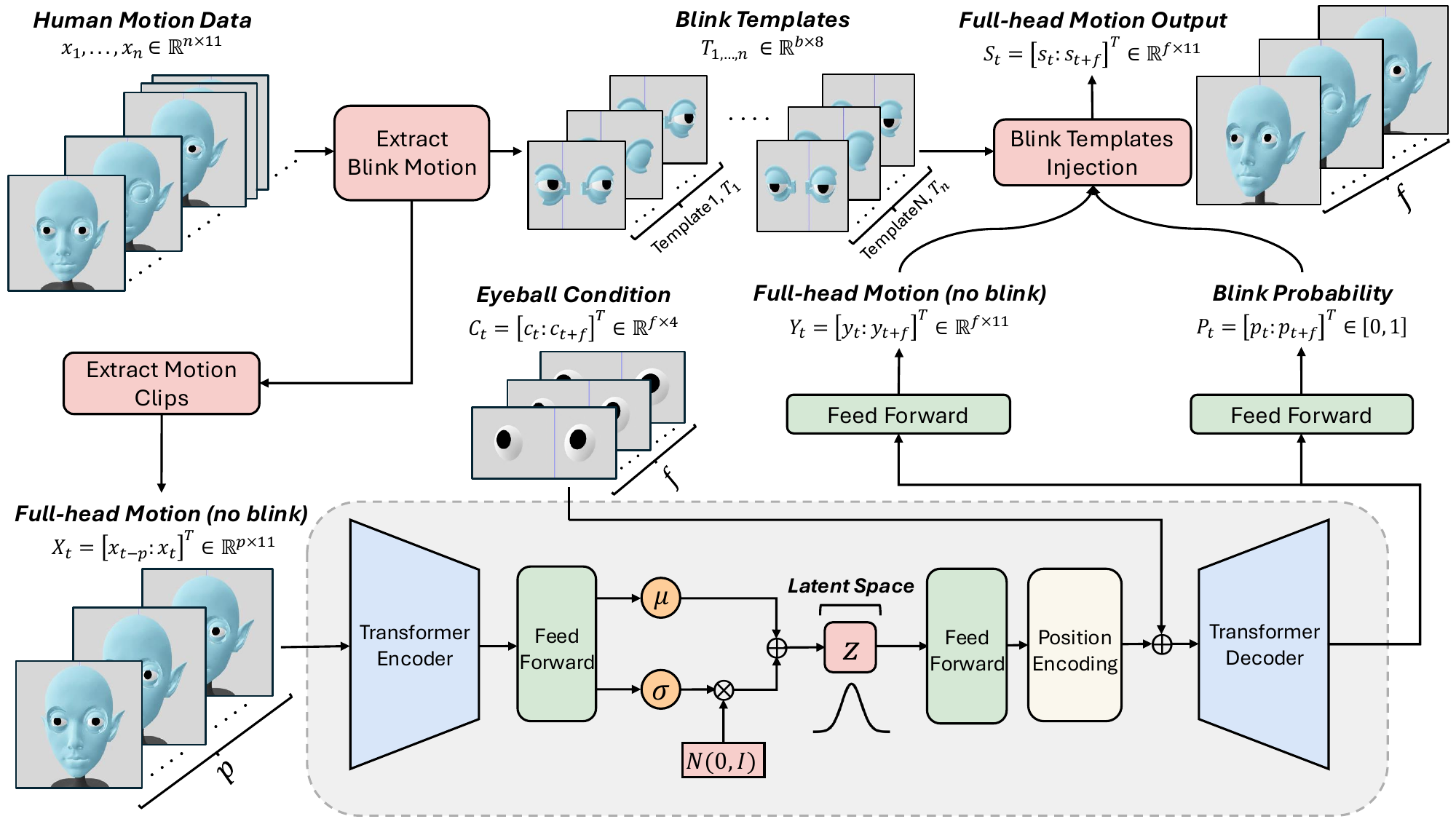}
    \caption{\textbf{Full pipeline of the Realistic Gaze Transformer (RGT).} 
    Top: Data preprocessing stage. Raw human motion sequences (head pose, eye gaze, and blinks) are segmented into clips. Blink events are detected and extracted as separate blink templates (small eyelid motion sequences), and those blink segments are removed from the sequences to produce blink-free training data. Bottom: Model architecture and inference. Given a past full-head motion sequence and a planned eyeball motion sequence, the Transformer-VAE model (encoder and decoder) predicts a future blink-free trajectory along with blink probabilities for each future frame. At inference time, when a blink is predicted (high probability), a corresponding blink template is injected into the generated trajectory, yielding the final full-head motion with realistic blinks (red frames show the injected blink).}
    \label{fig:rgt_arch}
    \vspace{-10pt}
\end{figure}

\textbf{Representation Learning for Motion.} Contrastive methods \cite{chen2020simple,tevet2022motionclip} and Temporal Convolutional Networks \cite{bai2018empirical} extract robust spatiotemporal embeddings that generalize to downstream motion tasks. Generative approaches advance this further: VAE and Transformer-VAE frameworks \cite{petrovich2021action,yuan2020dlow, vaswani2017attention,kingma2014vae}, and more recently diffusion models \cite{yan2023gazemodiff,chi2023diffusion,tevet2022human}, enable controllable and physically plausible motion synthesis, ranging from gaze-conditioned prediction to text-conditioned human motion generation.

\section{Methods}
\label{sec:method}
We address the problem of generating realistic, expressive gaze behavior for a
humanoid face robot. The platform is the soft-skinned humanoid face introduced by
Hu~\cite{hu2026lipmotions}, whose actuated facial mechanism we repurpose for
gaze: in our configuration, 11 degrees of freedom (DoF) govern neck, eyeball,
and eyelid motion, driven in real time (see Appendix~\ref{sec:hardware_control}). Central to our formulation is a decoupling
of high-level cognitive reasoning from low-level motor control: an upstream
Vision-Language Model determines where to attend, producing a target gaze
trajectory, while the learned action head is responsible for how that target
is realized as lifelike full-head motion. Given a dataset of human gaze motion sequences, each training sample consists of a history of full-head motion
$X_t \in \mathbb{R}^{p \times 11}$ over $p$ past frames (neck, eyeballs,
eyelids) and a future eyeball trajectory $C_t \in \mathbb{R}^{f \times 4}$
over $f$ future frames. We learn a Transformer-VAE in which an encoder
$f_\theta$ maps $X_t$ to a latent code $z \in \mathbb{R}^{64}$ parameterizing
the posterior $q_\theta(z \mid X_t)$, and a decoder $g_\phi$ jointly predicts
the future full-head motion $\hat{Y}_t \in \mathbb{R}^{f \times 11}$ and
per-frame blink probabilities $\hat{P}_t \in [0,1]^{f}$ from $z$ and $C_t$: $\hat{Y}_t,\; \hat{P}_t \;=\; g_\phi\!\left(f_\theta(X_t),\, C_t\right)$.
The objective is to track $C_t$ while preserving the temporal structure of
the human motion prior encoded in $X_t$. Predicted sequences
are translated into motor commands that drive the robot's head, eyes, and
eyelids (Appendix~\ref{sec:hardware_control}). The remainder of this section formalizes each component: the targeting
pipeline that produces $C_t$ (Sec.~\ref{sec:vlm_targeting}), the data and
conditioning used to train the action head (Sec.~\ref{sec:preprocessing},
Sec.~\ref{sec:eye_cond}), and the Realistic Gaze Transformer with its training
procedure (Sec.~\ref{sec:rgt}, Sec.~\ref{sec:training}).

\subsection{VLM-Guided Gaze Targeting}
\label{sec:vlm_targeting}

To drive the action head, we need a target eyeball trajectory
$C_t \in \mathbb{R}^{f \times 4}$ that specifies where the robot should look.
This trajectory is produced by a two-stage pipeline inspired by human attention.
The first stage is a deliberative module built on a Vision-Language Model
(we adopt \texttt{Gemma 3}~\cite{gemma3_2025}, though the design is agnostic to
the specific VLM): given a scene image $I_t$ and a natural-language instruction
$\ell$ (e.g., \emph{``Notice the man in the white shirt''}), prompt engineering
constrains its output to a structured referent
$r_t = \mathrm{VLM}(I_t, \ell)$ describing the attended target and a rough region
of interest. The second stage is a reactive perception module
(we use FastSAM~\cite{zhao2023fastsam}) that refines $r_t$ into a frame-rate
fixation $u_t \in \mathbb{R}^{2}$, maintaining the target between successive VLM
calls. We project $u_t$ into the robot's eyeball frame to obtain an instantaneous
gaze direction, which the eyeball trajectory generator unrolls into the
conditioning signal $C_t$ for the RGT (Sec.~\ref{sec:eye_cond}). Prompt templates
and model-selection details are provided in
Appendix~\ref{sec:Visual Perception Module}.
 
\subsection{Data Preprocessing}
\label{sec:preprocessing}
 
The markerless, webcam-based facial motion data (see Appendix~\ref{sec:dataset_description}) comprises
timestamped sequences with 11-dimensional features per frame, including head
orientation (yaw, pitch, roll), binocular eye gaze directions (yaw and pitch for
each eye), and upper and lower eyelid motions (pitch for each eyelid). First, abrupt
spikes in the eye gaze channels are identified via threshold-based peak detection and
corrected by interpolating across the affected intervals. Second, high-frequency
noise is reduced using signal-specific low-pass filtering techniques.
Blink labels $P_t$ are extracted from the upper eyelid signal using inverted peak
detection. We generate both binary and soft labels, where the latter are obtained by
convolving discrete blink events with a Gaussian kernel to provide smooth temporal
supervision during training. All channels are resampled to a 60-Hz
robot-command grid before denoising; eye and eyelid artifacts are handled
separately (Appendix~\ref{sec:Preprocessing} and~\ref{sec:blink}). The trajectories
are retargeted to robot coordinates, including approximately $\pm30^\circ$ eye
yaw and $\pm20^\circ$ eye pitch. They are a shared kinematic reference, not
optical-motion-capture ground truth.

\subsection{Eyeball-Trajectory Generator}
\label{sec:eye_cond}
 
At both training and test time, we require a 4-D eyeball-condition signal
$C_t = [c_t;\, c_{t+1}; \dots]^{\!\top} \in \mathbb{R}^{f \times 4}$ that specifies
where the robot should look. When the processed human trajectories are available (pre-training and
fine-tuning) we take the ground-truth binocular gaze directions
$(\text{yaw}^{\mathrm{R}}, \text{pitch}^{\mathrm{R}}, \text{yaw}^{\mathrm{L}},
\text{pitch}^{\mathrm{L}})$ as $C_t$. During the evaluation of the real robot, \(C_t\) is synthesized online by the condition generator in Appendix~\ref{sec:eye_cond_online}.
This procedure provides task-directed eye commands while the RGT predicts
coordinated motion and blink timing; the auxiliary template path realizes
blink eyelid playback (Appendix~\ref{sec:blink}). Optional stochastic eye-command
perturbations are applied only to the procedural condition stream at deployment,
not to the ground-truth conditions used in Table~\ref{tab:overall_joint_mse}.

\subsection{Realistic Gaze Transformer (RGT)}
\label{sec:rgt}

Our objective is to synthesize human-like gaze that remains controllable by a
task-level eyeball trajectory while preserving the rich micro-dynamics of natural
head--eye coordination. To this end, we introduce a Transformer-VAE
(Fig.~\ref{fig:rgt_arch}) whose self-attention layers capture the delayed,
multi-scale coupling between eyes, head, and eyelids. The motion history $X_t$
is compressed into a latent vector $z$ that parameterizes a low-dimensional
manifold of plausible human gaze dynamics. Because the decoder has no skip
connection to $X_t$, all information about the recent past must pass through
$z$, forcing it to summarize both kinematic state---inertia, head overshoot,
microsaccadic patterns---and motion style. At inference, conditioning on $z$
biases generation toward this manifold. Stochastic sampling of $z$ yields
different realizations of the same target $C_t$. Since $C_t$ enters after the latent
bottleneck, the model cleanly separates \emph{where} the robot looks ($C_t$)
from \emph{how} the movement unfolds ($z$). We measure the resulting rollout
diversity in the bottleneck ablation rather than treating it as a perceptual score.

\paragraph{Model Architecture}
Figure~\ref{fig:rgt_arch} depicts the Realistic Gaze Transformer (RGT), a
Transformer-VAE comprising three functional blocks: (i) a motion-history
encoder, (ii) a latent-conditioned decoder, and (iii) an auxiliary blink head.
Each training sample pairs a fixed-length history window
$X_t \in \mathbb{R}^{p \times 11}$ (neck, eyeballs, eyelids) with a future
eyeball trajectory $C_t \in \mathbb{R}^{f \times 4}$. A linear layer projects
the 11-D face vector to a $d_{\text{model}} = 128$ embedding, followed by a
sine--cosine positional encoding; the same procedure, with a dedicated weight
matrix, is applied to the 4-D eye condition. A 3-layer Transformer encoder with
$d_{\text{ff}} = 512$ processes the embedded history---we found this depth
sufficient to capture eye--head coupling on the limited
markerless recording set. The resulting sequence
$\mathbf{h} \in \mathbb{R}^{p \times 128}$ is flattened and passed through a
small aggregation MLP. With $p=60$, all ordered history tokens contribute to
its 7,680-dimensional input; the representation retains their temporal positions
instead of reading only the final frame. Two linear heads then produce the posterior
parameters $\mu, \log\sigma^{2} \in \mathbb{R}^{64}$. We then sample
$z \in \mathbb{R}^{64}$ via the reparameterization trick, linearly expand it to
$\mathbf{z}_{1:f} \in \mathbb{R}^{f \times 128}$, and add it element-wise to the
eye-condition embeddings, injecting style information at every decoder position.
A 4-layer Transformer decoder with $d_{\text{ff}} = 512$ processes the combined
tokens, after which two task-specific heads predict future facial motion
$\hat{Y}_t \in \mathbb{R}^{f \times 11}$ and blink probabilities
$\hat{P}_t \in [0,1]^{f}$ (MLP with ReLU and Sigmoid activations).

\subsection{Model Training Procedure}
\label{sec:training}

We train the Realistic Gaze Transformer (RGT) on S1 in two stages: (1) self-supervised
pre-training, which learns a compact manifold of natural head--eye dynamics, and
(2) autoregressive fine-tuning, which adds long-horizon roll-out stability while
tracking task-level eyeball commands. To eliminate temporal leakage, each raw
sequence is first split chronologically (80\% train, 20\% validation), and only
then are stride-1 sliding windows extracted within each partition.
 
\paragraph{Stage 1 -- Self-supervised pre-training.}
We construct overlapping windows with a 60-frame (one-second) history and a
future block of $f$ frames from the normalized markerless data. Each sample
comprises (i) a history segment $X_t \in \mathbb{R}^{p \times 11}$ with $p=60$
and (ii) a future eyeball trajectory $C_t \in \mathbb{R}^{f \times 4}$.
In the reported benchmark, $f=3$ per model call. The Transformer encoder embeds $X_t$
and produces a latent posterior; a latent code $z \sim \mathcal{N}(\mu, \sigma^{2})$
is added to the decoder input tokens together with $C_t$, yielding reconstructions
$\hat{Y}_t \in \mathbb{R}^{f \times 11}$ and blink probabilities
$\hat{P}_t \in [0,1]^{f}$. Training minimizes
\begin{equation}
\mathcal{L}_{\text{pre}} \;=\;
\operatorname{MSE}(Y_t, \hat{Y}_t)
\;+\;
\lambda_{\text{KL}}\,
\operatorname{KL}\!\big(q_\theta(z \mid X_t)\,\|\,\mathcal{N}(0, I)\big)
\;+\;
\lambda_{\text{BCE}}\,
\operatorname{BCE}(P_t, \hat{P}_t),
\end{equation}
with $\lambda_{\text{KL}} = 10^{-4}$ and $\lambda_{\text{BCE}} = 0.1$. We use Adam \cite{kingma2015adam}
($\mathrm{lr} = 10^{-3}$, batch size $= 128$) for up to 400 epochs, employing a
10-epoch plateau scheduler ($\mathrm{lr} \times 0.5$) and 20-epoch early stopping on
the validation loss. The best checkpoint is used to initialize Stage 2.
 
\paragraph{Stage 2 -- Autoregressive fine-tuning.}
To expose the RGT to accumulating prediction errors, we build longer segments of
length $p + f \cdot m$ frames, where $m$ is the maximum autoregressive roll-out
horizon. During each update, we feed the seed history $X_t$ into the encoder and
unroll the decoder for $m$ prediction blocks of $f$ frames each; at step $k$
($0 \le k < m$) the decoder is conditioned on the eye commands $C_{t+k}$ and the
model's own previous predictions. We accumulate the same VAE${+}$BCE loss per block
and back-propagate their mean. We fine-tune with a $10\times$ smaller learning rate
($\mathrm{lr} = 10^{-4}$) to preserve the pre-trained latent prior, and again monitor
a held-out validation set with the same early-stopping policy. 

\begin{figure}[t]
    \centering
    \includegraphics[width=1.0\linewidth]{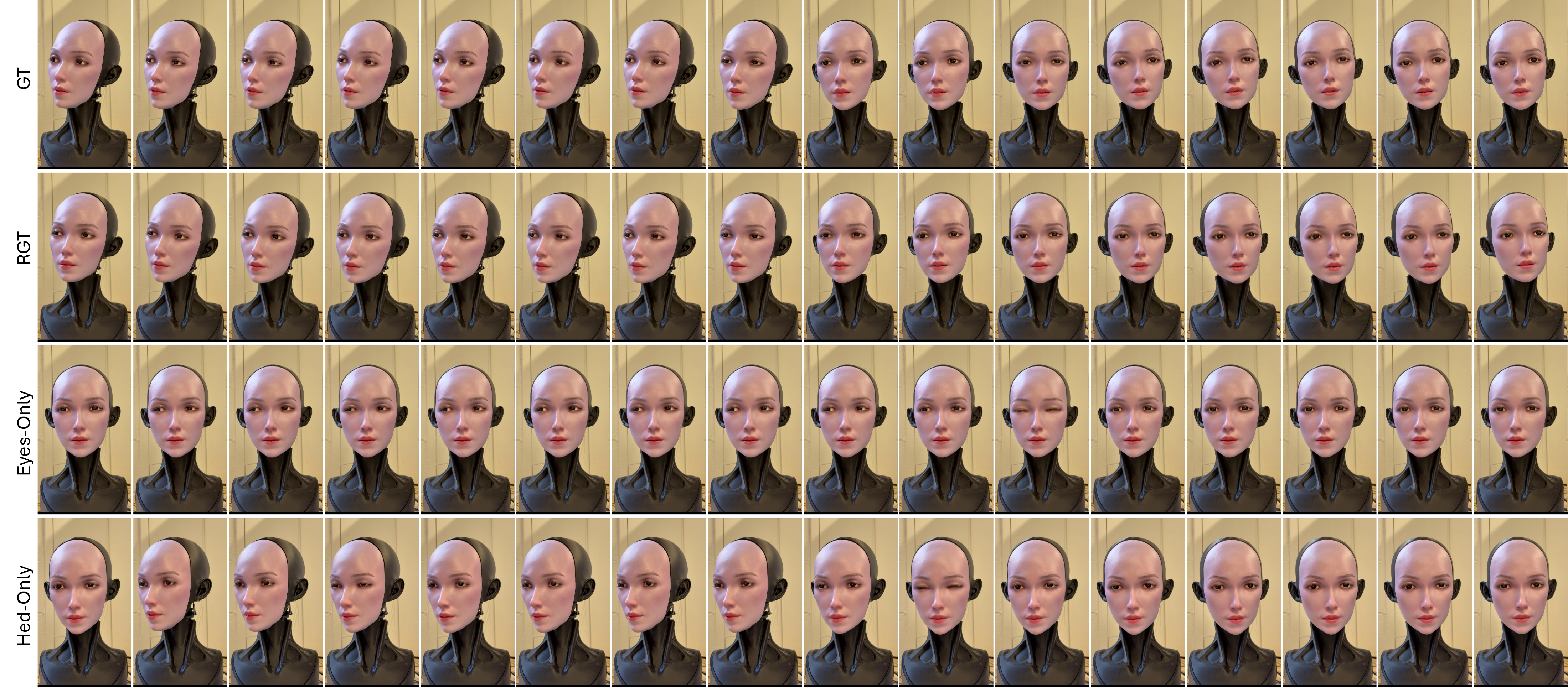}
    \caption{\textbf{Qualitative comparison of gaze behaviors across controllers.}
    Each row shows a time sequence of robot face images during a representative
    leftward gaze shift. \textbf{Top}: Ground-truth human motion (GT) exhibits
    coordinated eye--head motion, with eyes leading the head.
    \textbf{Second}: RGT reproduces the illustrated sequence's eye--head timing.
    \textbf{Third}: Eyes-Only
    rotates the eyes to mechanical extremes while keeping the head fixed, resulting
    in strained, unnatural motion. \textbf{Bottom}: Head-Only turns the head abruptly
    toward the target while leaving the eyes static, exaggerating head movement and
    yielding sluggish gaze redirection. Each column represents 1/3 second of motion.}
    \label{fig:motion_visualization}
    \vspace{-10pt}
\end{figure}

\section{Experimental Results}
\label{sec:result}

We evaluate RGT against two families of baselines using S1 alone. The first comprises three
heuristic, rule-based controllers: Eyes-Only rotates the eyeballs toward the target
with the head fixed; Head-Only turns the neck toward the target with the eyes
centered; Eye--Head Coord.\ drives both channels with the eyes leading, following
fixed coordination and blink heuristics. All three run in closed loop. Because
Eyes-Only and Head-Only are penalized for the joints they leave pinned, we
additionally report each controller's error over only its actively commanded joints
(``w/o fixed''), isolating tracking quality from the cost of unmoved joints. The
second family comprises learned baselines that isolate the contribution of our
architecture: RNN \cite{hochreiter1997lstm} replaces self-attention with recurrence; VAE-CNN replaces the
Transformer with ResNet \cite{he2016resnet} encoders/decoders; Flow Matching \cite{lipman2023flow} replaces our
latent-variable VAE with a continuous-flow generative objective. All learned
baselines are matched to RGT-Base in parameter count and trained under an identical
protocol; RGT-Small and RGT-Large are reported separately as scale ablations.
Every controller---rule-based or learned---receives the same four-dimensional
ground-truth eye condition. The benchmark therefore isolates motion generation
from errors in VLM referent selection and visual tracking. From chronologically
held-out S1 tails, we extract 1,848 clips with 60 history frames and a
120-frame (2-s) future at stride 10. A learned model predicts three frames per
call, feeds back the first frame as history, and yields 117 scored updates over
the seven neck and binocular-eye channels. The separate eyelid and blink path
is not included in this score. RGT and VAE-CNN draw one reparameterized latent
per call; neither uses a posterior mean or best-of-$K$ selection. For $N$ clips,
$T=117$ scored updates, and $J=7$ joints, the reported joint-angle MSE is
$\frac{1}{NTJ}\sum_{i=1}^{N}\sum_{t=1}^{T}\sum_{j=1}^{J}
(\hat y_{itj}-y_{itj})^2$ in $\mathrm{rad}^2$.
Table~\ref{tab:overall_joint_mse} contains one stochastic evaluation run;
evaluation seeds and confidence intervals were not recorded for that run.

\subsection{Quantitative and Qualitative Results}

Figure~\ref{fig:motion_visualization} compares ground-truth motion, RGT,
Eyes-Only, and Head-Only during a representative leftward gaze shift. In the
recorded sequence the eyes move before the head, and the RGT frames follow the
same broad temporal ordering. Eyes-Only reaches a larger ocular rotation while
the head remains fixed; Head-Only redirects the neck while the eyes remain near
their initial orientation. These distinct mechanisms explain what the visual
comparison is intended to show, but the frames are one selected example rather
than a distribution of outcomes. No controlled observer ratings were collected,
so they cannot establish which motion people perceive as more natural.

Table~\ref{tab:overall_joint_mse} reports joint-angle MSE for the six S1 behavior
sequences in Table~\ref{tab:motion_summary}. The learned models have lower
reported seven-channel error than all three heuristic controllers. The gap is
substantial even when the fixed-joint contribution is removed from each
heuristic's diagnostic ``w/o fixed'' score, although those rows use fewer
channels and should not be compared as an identical metric. Among the learned
models, RGT-Base has the lowest overall MSE. Its unrounded value is
$2.19\times10^{-4}\,\mathrm{rad}^2$, versus
$3.04\times10^{-4}\,\mathrm{rad}^2$ for RNN, a descriptive 27.9\% reduction.
RGT-Base leads on BIDs 2, 3, and 5; RGT-Large leads on BIDs 1 and 4, and RNN
on BID 6. The behavior-wise variation matters: the aggregate result does not
imply that the Base configuration is best for every motion pattern.

The single-step sliding-window evaluation in Appendix~\ref{app:approach1}
provides a complementary view of the same model family. Flow Matching has the
lowest error under that protocol, yet its closed-loop error is higher in
Table~\ref{tab:overall_joint_mse}, where generated actions become part of the
next history. This reversal is consistent with errors accumulating when the
input distribution shifts away from recorded histories. Autoregressive
fine-tuning explicitly trains RGT on such self-generated histories, which is
also reflected in the training ablation of Fig.~\ref{fig:ar_training_comparison}.
The results motivate evaluating rollout behavior rather than selecting a model
only from teacher-forced reconstruction.

Runtime determines whether the controller can update within a 60-Hz cycle.
Measured on an NVIDIA RTX 2080 Ti, RGT-Base inference is below the 16.7-ms
frame interval, while the VLM operates asynchronously and does not need to
produce a new referent at every control step. The reported latency applies to
RGT inference, not to the entire perception, communication, and actuation
pipeline. Likewise, the one-run MSE difference and the qualitative frames do
not establish statistical significance or perceived naturalness; repeated
stochastic evaluations and an observer study would address those questions.

\begin{table}
    \centering
    \caption{Closed-loop seven-joint MSE ($\mathrm{rad}^2$) over 1,848 S1 clips.
    A single stochastic sample is evaluated per model call. Values are from one
    run and are rounded; no uncertainty intervals are available. ``w/o fixed''
    rows score only active joints and are not directly comparable with the
    seven-joint rows.}
    \scriptsize
    \setlength{\tabcolsep}{12pt}
    \begin{tabular}{lccccccc}
        \toprule
        \textbf{Method} & \textbf{BID 1} & \textbf{BID 2} & \textbf{BID 3} & \textbf{BID 4} & \textbf{BID 5} & \textbf{BID 6} & \textbf{Overall} \\
        \midrule
        Head-Only                      & 5.8e-2          & 5.9e-2          & 7.3e-2          & 7.2e-2          & 7.2e-2          & 6.8e-2          & 6.8e-2 \\
        Eye-Only                       & 5.1e-2          & 5.4e-2          & 5.7e-2          & 5.6e-2          & 5.5e-2          & 5.2e-2          & 5.2e-2 \\
        Eye-Head Coord.                & 1.3e-2          & 1.2e-2          & 3.9e-2          & 1.5e-2          & 1.7e-2          & 1.4e-2          & 1.8e-2 \\
        Head-Only (w/o fixed)          & 1.6e-2          & 1.3e-2          & 5.8e-2          & 2.3e-2          & 3.2e-2          & 2.1e-2          & 3.1e-2 \\
        Eye-Only (w/o fixed)           & 1.7e-2          & 1.8e-2          & 1.5e-2          & 2.3e-2          & 1.1e-2          & 1.6e-2          & 1.7e-2 \\
        \midrule
        RNN                            & 1.3e-4          & 2.0e-4          & 1.6e-4          & 2.2e-4          & 3.7e-4          & \textbf{9.8e-4} & 3.0e-4 \\
        VAE-CNN                        & 2.5e-4          & 2.6e-4          & 1.3e-4          & 1.8e-4          & 3.3e-4          & 1.7e-3          & 4.6e-4 \\
        Flow Matching (16-step)        & 5.5e-4          & 1.0e-3          & 1.2e-3          & 2.0e-3          & 1.3e-3          & 1.8e-3          & 1.1e-3 \\
        RGT-Base                       & 3.9e-5          & \textbf{6.9e-5} & \textbf{5.3e-5} & 9.5e-5          & \textbf{2.9e-4} & 1.0e-3          & \textbf{2.2e-4} \\
        \midrule
        RGT-Large                      & \textbf{3.6e-5} & 7.1e-5          & 6.3e-5          & \textbf{7.4e-5} & 3.4e-4          & 1.1e-3          & 2.4e-4 \\
        RGT-Small                      & 5.9e-5          & 1.1e-4          & 8.0e-5          & 1.1e-4          & 3.7e-4          & 1.1e-3          & 2.6e-4 \\
        \bottomrule
    \end{tabular}
    \label{tab:overall_joint_mse}
    \vspace{-10pt}
\end{table}

\subsection{Adaptation Across Different Human Subjects}
\label{sec:subject_adaptation}

We next test whether the S1-pretrained model can adapt to motion from other
people without repeating full pretraining. For each of Subjects~2--6, we use
only the training prefix of that subject's S2 sequence for fine-tuning and
evaluate on its corresponding held-out tail
(Appendix~\ref{sec:dataset_description}). These are separate subject-specific
adaptations, not one model jointly trained on all five held-out subjects.
Table~\ref{tab:s2_approach4_mse} reports the base and adapted joint-angle MSE
for each person. The reported overall error decreases from
$1.8\times10^{-3}$ to $3.6\times10^{-4}\,\mathrm{rad}^2$, approximately a
fivefold reduction, and all five subject columns improve. The magnitude of
the gain differs by subject, so the aggregate does not describe every
individual equally. As a separate qualitative check, adapted models were
also run with the same procedurally generated eye targets, which were not used
in training or validation. The resulting visualizations permit inspection of
eye--head balance, blink frequency, and head-motion range, but these style
attributes have no independent quantitative or observer rating here. The
measured result supports adaptation in robot joint space across the five
sampled subjects; the small, non-clinical cohort does not establish
performance across broader populations or atypical ocular kinematics.

\subsection{Ablation Studies}
We conduct ablation studies to better understand the contributions of key components in our training framework, focusing on three factors: (1) the latent bottleneck design, comparing the stochastic VAE bottleneck against a deterministic autoencoder and a skip-connected variant that bypasses the bottleneck; (2) the two-stage training schedule, isolating the effects of teacher-forced pre-training and autoregressive fine-tuning; and (3) the encoder/decoder depth, scaling the transformer backbone from the base configuration up to larger variants.

\begin{figure}[t]
    \centering
    \begin{minipage}[c]{0.58\linewidth}
        \centering
        \captionof{table}{Joint-angle MSE ($\mathrm{rad}^2$) on the held-out tail of each S2 subject sequence.}
        \scriptsize
        \setlength{\tabcolsep}{3pt}
        \begin{tabular}{lcccccc}
            \toprule
            \textbf{Method} & \textbf{Subj 2} & \textbf{Subj 3} & \textbf{Subj 4} & \textbf{Subj 5} & \textbf{Subj 6} & \textbf{Overall} \\
            \midrule
            Base Model  & 1.2e-3 & 1.5e-3 & 3.7e-3 & 1.1e-3 & 1.3e-3 & 1.8e-3 \\
            Fine-Tuned  & \textbf{3.7e-4} & \textbf{3.2e-4} & \textbf{7.3e-4} & \textbf{1.5e-4} & \textbf{2.4e-4} & \textbf{3.6e-4} \\
            \bottomrule
        \end{tabular}
        \label{tab:s2_approach4_mse}

        \vspace{8pt}

        \captionof{table}{Latent bottleneck ablation; APD uses $K=10$ rollouts.}
        \scriptsize
        \setlength{\tabcolsep}{4pt}
        \begin{tabular}{lccc}
            \toprule
            \textbf{Variant} & \textbf{MSE} ($\mathrm{rad}^2$) & \textbf{APD} ($\mathrm{rad}^2$) & \textbf{Std} ($\mathrm{rad}$) \\
            \midrule
            Deterministic AE        & \textbf{2.3e-4} & 0.0 (collapsed) & 0.0 \\
            VAE + skip connection   & 2.4e-4          & 5.5e-7          & 4.4e-4 \\
            VAE (ours)              & 2.6e-4          & \textbf{4.4e-6} & 1.2e-3 \\
            \bottomrule
        \end{tabular}
        \label{tab:latent_bottleneck}
    \end{minipage}
    \hfill
    \begin{minipage}[c]{0.40\linewidth}
        \centering
        \includegraphics[width=\linewidth]{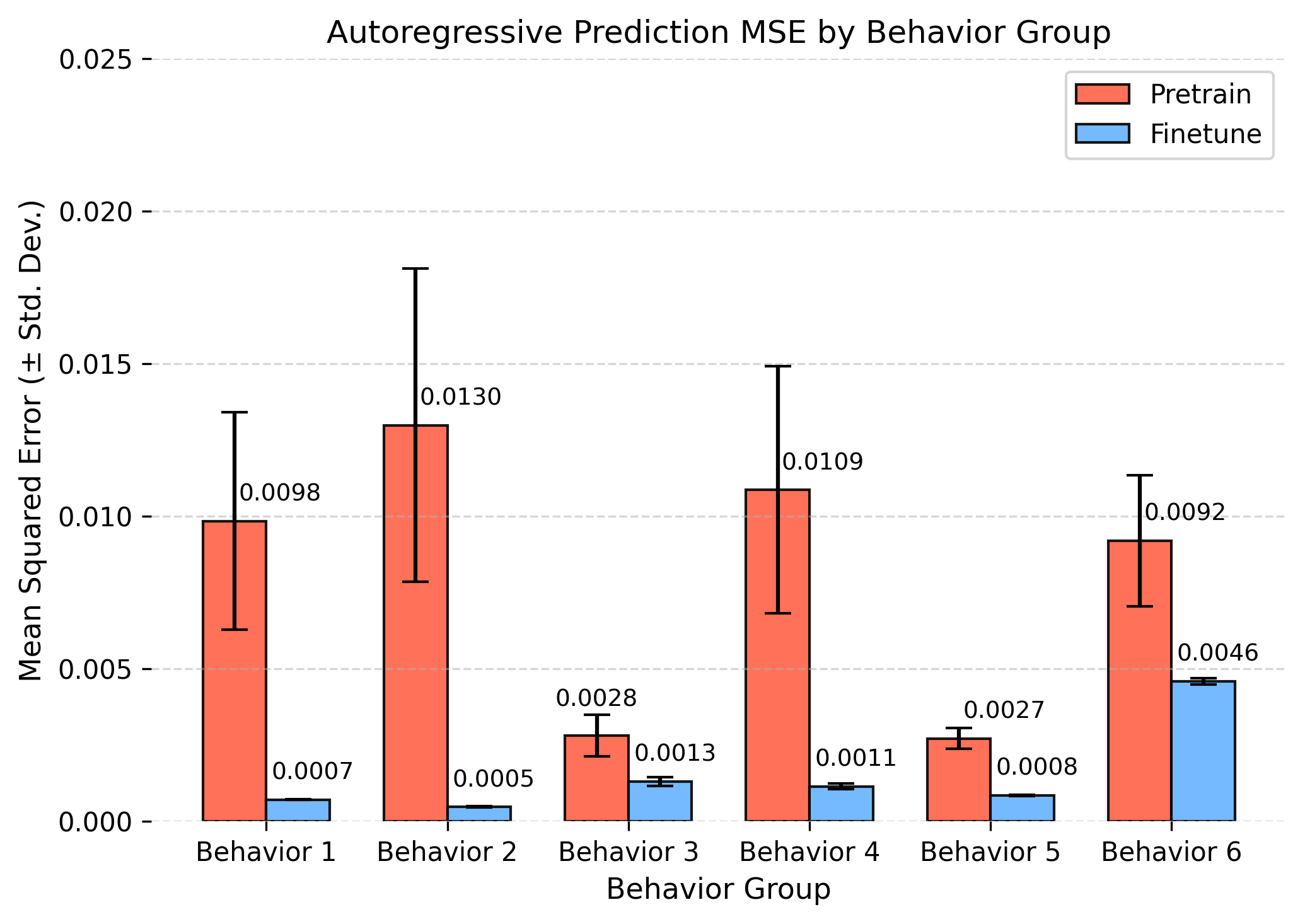}
        \captionof{figure}{\textbf{Training Ablation Studies} Comparison of pretraining and finetuning across different BIDs}
        \label{fig:ar_training_comparison}
    \end{minipage}
    \vspace{-10pt}
\end{figure}

We first compare the full RGT against a deterministic autoencoder ($z{=}\mu$, KL
disabled) and a skip-connected variant whose decoder cross-attends to the encoder
outputs, bypassing the bottleneck. Table~\ref{tab:latent_bottleneck} reports
reconstruction MSE under closed-loop AR rollout alongside the average pairwise
distance (APD) between $K{=}10$ stochastic rollouts of the same clip. APD is
reported in $\mathrm{rad}^2$ in Table~\ref{tab:latent_bottleneck};
it measures output diversity rather than fidelity or observer preference.
The deterministic AE yields one rollout ($\mathrm{APD}{=}0$), and the skip
variant's APD is smaller than the no-skip VAE's. The no-skip VAE has APD
$4.4\times10^{-6}\,\mathrm{rad}^2$, compared with
$5.5\times10^{-7}\,\mathrm{rad}^2$ for the skip model, while their
reconstruction MSEs remain of the same order. This is consistent with the
decoder relying less on the latent code when it can bypass the bottleneck, but
APD alone does not diagnose posterior collapse or show that the different
rollouts are perceptually plausible. The deterministic variant also obtains a
slightly lower reconstruction error, making the fidelity--diversity tradeoff
visible rather than treating diversity as a free gain. Next, comparing pretraining-only against
models further fine-tuned with AR training
(Fig.~\ref{fig:ar_training_comparison}), the fine-tuned models consistently
achieve lower prediction errors across all behavior groups, confirming that AR
fine-tuning is critical for reducing compounding errors and improving long-term
motion stability. Finally, we compare three RGT depths under the same recipe:
Small (2 enc, 3 dec, $d_{\text{ff}}{=}256$), Base (3 enc, 4 dec,
$d_{\text{ff}}{=}512$), and Large (4 enc, 6 dec, $d_{\text{ff}}{=}512$). The last
three rows of Table~\ref{tab:overall_joint_mse} show that Base has lower overall
MSE than Small or Large in this run; the table alone does not identify the cause
of the larger model's difference. We therefore adopt Base as the default RGT.

\section{Conclusion}
\label{sec:conclusion}

We presented a gaze-control framework that separates semantic target selection
from the generation of eye, head, and eyelid motion on a soft-skinned humanoid
face. An asynchronous VLM and a reactive perception module provide the target;
the Realistic Gaze Transformer uses a one-second motion history and a planned
binocular-eye trajectory to generate coordinated 11-DoF commands in real time.
Its Transformer-VAE learns from retargeted human recordings, and autoregressive
fine-tuning addresses the distribution shift caused by feeding predictions
back into the controller. Blink timing is predicted by an auxiliary head and
realized through template playback. In the reported ground-truth-target
benchmark, RGT-Base has the lowest overall seven-joint MSE among the tested
controllers. Bottleneck and training ablations, together with adaptation on
five further subjects, provide complementary evidence about the motion model.
The physical demonstration shows the targeting and control components working
together. These findings motivate a next HRI evaluation that tests whether
people can infer the robot's attention during handover or collaborative search,
as well as whether they prefer its gaze to alternative controllers.

\section{Limitations and Future Work}

Several limitations bound the interpretation of these results. The primary
benchmark conditions every controller on the same ground-truth eye target,
which isolates how the robot moves but does not measure whether the VLM picks
the intended referent or whether the reactive module maintains a fixation
under occlusion. A controlled target-perturbation sweep and labeled
open-vocabulary grounding cases are needed to assess those errors and their
effect on the complete pipeline. The VLM also operates below the control rate
and supplies no explicit policy for when to shift gaze during turn-taking or
mutual-gaze maintenance. Audio and conversational cues, together with feedback
from the human partner's gaze and expression, could support such decisions;
the present system does not adapt to the partner's affect in closed loop.

The reported MSE table is one stochastic run without fixed evaluation seeds or
confidence intervals. Repeated runs with prespecified seeds would quantify
uncertainty, especially for the smaller gaps among learned models. MSE measures
robot-space kinematic agreement and APD measures stochastic diversity; neither
is an observer rating. Controlled studies are needed to assess perceived
naturalness, attention legibility, preference, and trust. The six-participant,
non-clinical sample and markerless webcam retargeting constrain generalization,
including to atypical ocular kinematics, different gaze conventions, and other
robot joint layouts. Per-eye recalibration and additional participant-specific
adaptation may help, but were not tested here. Blink templates simplify eyelid
playback and do not provide a directly interpretable control over blink style
or its relation to attention. Finally, the scored rollouts last two seconds,
leaving longer-interaction drift unmeasured. Broader deployment should also
study possible unwarranted trust or anthropomorphic projection.

\clearpage
\bibliography{camera_ready}

\begin{thebibliography}{48}
\providecommand{\natexlab}[1]{#1}
\providecommand{\url}[1]{\texttt{#1}}
\expandafter\ifx\csname urlstyle\endcsname\relax
  \providecommand{\doi}[1]{doi: #1}\else
  \providecommand{\doi}{doi: \begingroup \urlstyle{rm}\Url}\fi

\bibitem[Frischen et~al.(2007)Frischen, Bayliss, and Tipper]{frischen2007gaze}
A.~Frischen, A.~P. Bayliss, and S.~P. Tipper.
\newblock Gaze cueing of attention: visual attention, social cognition, and individual differences.
\newblock \emph{Psychological bulletin}, 133\penalty0 (4):\penalty0 694--724, 2007.

\bibitem[Stephenson et~al.(2021)Stephenson, Edwards, and Bayliss]{stephenson2021gaze}
L.~J. Stephenson, S.~G. Edwards, and A.~P. Bayliss.
\newblock From gaze perception to social cognition: The shared-attention system.
\newblock \emph{Perspectives on Psychological Science}, 16\penalty0 (3):\penalty0 553--576, 2021.

\bibitem[Kompatsiari et~al.(2017)Kompatsiari, Tikhanoff, Ciardo, Metta, and Wykowska]{kompatsiari2017importance}
K.~Kompatsiari, V.~Tikhanoff, F.~Ciardo, G.~Metta, and A.~Wykowska.
\newblock The importance of mutual gaze in human-robot interaction.
\newblock In \emph{Social Robotics: 9th International Conference, ICSR 2017, Tsukuba, Japan, November 22-24, 2017, Proceedings 9}, pages 443--452. Springer, 2017.

\bibitem[Pfeiffer et~al.(2013)Pfeiffer, Vogeley, and Schilbach]{pfeiffer2013gaze}
U.~J. Pfeiffer, K.~Vogeley, and L.~Schilbach.
\newblock From gaze cueing to dual eye-tracking: novel approaches to investigate the neural correlates of gaze in social interaction.
\newblock \emph{Neuroscience \& Biobehavioral Reviews}, 37\penalty0 (10):\penalty0 2516--2528, 2013.

\bibitem[Argyle and Dean(1965)]{argyle1965eye}
M.~Argyle and J.~Dean.
\newblock Eye-contact, distance and affiliation.
\newblock \emph{Sociometry}, 28\penalty0 (3):\penalty0 289--304, 1965.

\bibitem[Kendon(1967)]{kendon1967some}
A.~Kendon.
\newblock Some functions of gaze-direction in social interaction.
\newblock \emph{Acta psychologica}, 26:\penalty0 22--63, 1967.

\bibitem[Normoyle et~al.(2013)Normoyle, Badler, Fan, Badler, Cassol, and Musse]{10.1145/2522628.2522630}
A.~Normoyle, J.~B. Badler, T.~Fan, N.~I. Badler, V.~J. Cassol, and S.~R. Musse.
\newblock Evaluating perceived trust from procedurally animated gaze.
\newblock In \emph{Proceedings of Motion on Games}, MIG '13, pages 141--148, New York, NY, USA, 2013. Association for Computing Machinery.
\newblock ISBN 9781450325462.

\bibitem[Fong et~al.(2003)Fong, Nourbakhsh, and Dautenhahn]{fong2003survey}
T.~Fong, I.~Nourbakhsh, and K.~Dautenhahn.
\newblock A survey of socially interactive robots.
\newblock \emph{Robotics and autonomous systems}, 42\penalty0 (3-4):\penalty0 143--166, 2003.

\bibitem[Breazeal(2003)]{breazeal2003toward}
C.~Breazeal.
\newblock Toward sociable robots.
\newblock \emph{Robotics and autonomous systems}, 42\penalty0 (3-4):\penalty0 167--175, 2003.

\bibitem[Bartneck and Forlizzi(2004)]{bartneck2004design}
C.~Bartneck and J.~Forlizzi.
\newblock A design-centred framework for social human-robot interaction.
\newblock In \emph{RO-MAN 2004. 13th IEEE international workshop on robot and human interactive communication (IEEE Catalog No. 04TH8759)}, pages 591--594. IEEE, 2004.

\bibitem[Dautenhahn(2007)]{dautenhahn2007socially}
K.~Dautenhahn.
\newblock Socially intelligent robots: dimensions of human--robot interaction.
\newblock \emph{Philosophical transactions of the royal society B: Biological sciences}, 362\penalty0 (1480):\penalty0 679--704, 2007.

\bibitem[Hu et~al.(2024)Hu, Chen, Lin, Wang, Wang, Mehlman, and Lipson]{hu2024human}
Y.~Hu, B.~Chen, J.~Lin, Y.~Wang, Y.~Wang, C.~Mehlman, and H.~Lipson.
\newblock Human-robot facial coexpression.
\newblock \emph{Science Robotics}, 9\penalty0 (88):\penalty0 eadi4724, 2024.

\bibitem[Chen et~al.(2021)Chen, Hu, Li, Cummings, and Lipson]{chen2021smile}
B.~Chen, Y.~Hu, L.~Li, S.~Cummings, and H.~Lipson.
\newblock Smile like you mean it: Driving animatronic robotic face with learned models.
\newblock In \emph{2021 IEEE International Conference on Robotics and Automation (ICRA)}, pages 2739--2746. IEEE, 2021.

\bibitem[Admoni and Scassellati(2017)]{admoni2017social}
H.~Admoni and B.~Scassellati.
\newblock Social eye gaze in human-robot interaction: a review.
\newblock \emph{Journal of Human-Robot Interaction}, 6\penalty0 (1):\penalty0 25--63, 2017.

\bibitem[Mutlu et~al.(2012)Mutlu, Kanda, Forlizzi, Hodgins, and Ishiguro]{mutlu2012conversational}
B.~Mutlu, T.~Kanda, J.~Forlizzi, J.~Hodgins, and H.~Ishiguro.
\newblock Conversational gaze mechanisms for humanlike robots.
\newblock \emph{ACM Transactions on Interactive Intelligent Systems (TiiS)}, 1\penalty0 (2):\penalty0 1--33, 2012.

\bibitem[Admoni et~al.(2014)Admoni, Dragan, Srinivasa, and Scassellati]{admoni2014deliberate}
H.~Admoni, A.~Dragan, S.~S. Srinivasa, and B.~Scassellati.
\newblock Deliberate delays during robot-to-human handovers improve compliance with gaze communication.
\newblock In \emph{Proceedings of the 2014 ACM/IEEE international conference on Human-robot interaction}, pages 49--56, 2014.

\bibitem[Pan et~al.(2020)Pan, Choi, Kennedy, McIntosh, Zamora, Niemeyer, Kim, Wieland, and Christensen]{pan2020realistic}
M.~K. Pan, S.~Choi, J.~Kennedy, K.~McIntosh, D.~C. Zamora, G.~Niemeyer, J.~Kim, A.~Wieland, and D.~Christensen.
\newblock Realistic and interactive robot gaze.
\newblock In \emph{2020 IEEE/RSJ International Conference on Intelligent Robots and Systems (IROS)}, pages 11072--11078. IEEE, 2020.

\bibitem[Paetzel et~al.(2020)Paetzel, Perugia, and Castellano]{paetzel2020persistence}
M.~Paetzel, G.~Perugia, and G.~Castellano.
\newblock The persistence of first impressions: The effect of repeated interactions on the perception of a social robot.
\newblock In \emph{Proceedings of the 2020 ACM/IEEE international conference on human-robot interaction}, pages 73--82, 2020.

\bibitem[Z{\l}otowski et~al.(2015)Z{\l}otowski, Proudfoot, Yogeeswaran, and Bartneck]{zlotowski2015anthropomorphism}
J.~Z{\l}otowski, D.~Proudfoot, K.~Yogeeswaran, and C.~Bartneck.
\newblock Anthropomorphism: opportunities and challenges in human--robot interaction.
\newblock \emph{International journal of social robotics}, 7:\penalty0 347--360, 2015.

\bibitem[Faraj et~al.(2021)Faraj, Selamet, Morales, Torres, Hossain, Chen, and Lipson]{faraj2021facially}
Z.~Faraj, M.~Selamet, C.~Morales, P.~Torres, M.~Hossain, B.~Chen, and H.~Lipson.
\newblock Facially expressive humanoid robotic face.
\newblock \emph{HardwareX}, 9:\penalty0 e00117, 2021.

\bibitem[Rawal et~al.(2022)Rawal, Koert, Turan, Kersting, Peters, and Stock-Homburg]{rawal2022exgennet}
N.~Rawal, D.~Koert, C.~Turan, K.~Kersting, J.~Peters, and R.~Stock-Homburg.
\newblock Exgennet: Learning to generate robotic facial expression using facial expression recognition.
\newblock \emph{Frontiers in Robotics and AI}, 8:\penalty0 730317, 2022.

\bibitem[Yang et~al.(2025)Yang, Liu, Sato, Minato, Liu, and Nishida]{yang2025hapi}
D.~Yang, Q.~Liu, W.~Sato, T.~Minato, C.~Liu, and S.~Nishida.
\newblock {HAPI}: A model for learning robot facial expressions from human preferences.
\newblock In \emph{2025 IEEE/RSJ International Conference on Intelligent Robots and Systems (IROS)}. IEEE, 2025.

\bibitem[Hu et~al.(2026)Hu, Lin, Goldfeder, Wyder, Cao, Tian, Wang, Wang, Wang, Zeng, Mehlman, Wang, Zeng, Chen, and Lipson]{hu2026lipmotions}
Y.~Hu, J.~Lin, J.~A. Goldfeder, P.~M. Wyder, Y.~Cao, S.~Tian, Y.~Wang, J.~Wang, M.~Wang, J.~Zeng, C.~Mehlman, Y.~Wang, D.~Zeng, B.~Chen, and H.~Lipson.
\newblock Learning realistic lip motions for humanoid face robots.
\newblock \emph{Science Robotics}, 11\penalty0 (110):\penalty0 eadx3017, 2026.

\bibitem[Roncone et~al.(2016)Roncone, Pattacini, Metta, and Natale]{roncone2016cartesian}
A.~Roncone, U.~Pattacini, G.~Metta, and L.~Natale.
\newblock A cartesian 6-{DoF} gaze controller for humanoid robots.
\newblock In \emph{Proceedings of Robotics: Science and Systems (RSS)}, Ann Arbor, Michigan, 2016.
\newblock \doi{10.15607/RSS.2016.XII.022}.

\bibitem[Mishra and Skantze(2022)]{mishra2022knowing}
C.~Mishra and G.~Skantze.
\newblock Knowing where to look: A planning-based architecture to automate the gaze behavior of social robots.
\newblock In \emph{2022 31st IEEE international conference on robot and human interactive communication (RO-MAN)}, pages 1201--1208. IEEE, 2022.

\bibitem[Moon et~al.(2014)Moon, Zheng, Troniak, Blumer, Gleeson, MacLean, Pan, and Croft]{moon2014meet}
A.~Moon, M.~Zheng, D.~M. Troniak, B.~A. Blumer, B.~Gleeson, K.~MacLean, M.~K. Pan, and E.~A. Croft.
\newblock Meet me where i'm gazing: how shared attention gaze affects human-robot handover timing.
\newblock In \emph{Proceedings of the 2014 ACM/IEEE international conference on Human-robot interaction}, pages 334--341, 2014.

\bibitem[Saran et~al.(2018)Saran, Majumdar, Short, Thomaz, and Niekum]{saran2018human}
A.~Saran, S.~Majumdar, E.~S. Short, A.~Thomaz, and S.~Niekum.
\newblock Human gaze following for human-robot interaction.
\newblock In \emph{2018 IEEE/RSJ International Conference on Intelligent Robots and Systems (IROS)}, pages 8615--8621. IEEE, 2018.

\bibitem[Chen et~al.(2020)Chen, Kornblith, Norouzi, and Hinton]{chen2020simple}
T.~Chen, S.~Kornblith, M.~Norouzi, and G.~Hinton.
\newblock A simple framework for contrastive learning of visual representations.
\newblock In \emph{International conference on machine learning}, pages 1597--1607. PMLR, 2020.

\bibitem[Tevet et~al.(2022)Tevet, Gordon, Hertz, Bermano, and Cohen-Or]{tevet2022motionclip}
G.~Tevet, B.~Gordon, A.~Hertz, A.~H. Bermano, and D.~Cohen-Or.
\newblock Motionclip: Exposing human motion generation to clip space.
\newblock In \emph{European Conference on Computer Vision}, pages 358--374. Springer, 2022.

\bibitem[Bai et~al.(2018)Bai, Kolter, and Koltun]{bai2018empirical}
S.~Bai, J.~Z. Kolter, and V.~Koltun.
\newblock An empirical evaluation of generic convolutional and recurrent networks for sequence modeling.
\newblock \emph{arXiv preprint arXiv:1803.01271}, 2018.

\bibitem[Petrovich et~al.(2021)Petrovich, Black, and Varol]{petrovich2021action}
M.~Petrovich, M.~J. Black, and G.~Varol.
\newblock Action-conditioned 3d human motion synthesis with transformer vae.
\newblock In \emph{Proceedings of the IEEE/CVF International Conference on Computer Vision}, pages 10965--10975, 2021.

\bibitem[Yuan and Kitani(2020)]{yuan2020dlow}
Y.~Yuan and K.~M. Kitani.
\newblock Dlow: Diversifying latent flows for diverse human motion prediction.
\newblock In \emph{European Conference on Computer Vision (ECCV)}, pages 346--364. Springer, 2020.

\bibitem[Vaswani et~al.(2017)Vaswani, Shazeer, Parmar, Uszkoreit, Jones, Gomez, Kaiser, and Polosukhin]{vaswani2017attention}
A.~Vaswani, N.~Shazeer, N.~Parmar, J.~Uszkoreit, L.~Jones, A.~N. Gomez, {\L}.~Kaiser, and I.~Polosukhin.
\newblock Attention is all you need.
\newblock In \emph{Advances in Neural Information Processing Systems}, pages 5998--6008, 2017.

\bibitem[Kingma and Welling(2014)]{kingma2014vae}
D.~P. Kingma and M.~Welling.
\newblock Auto-encoding variational {B}ayes.
\newblock In \emph{International Conference on Learning Representations}, 2014.

\bibitem[Yan et~al.(2023)Yan, Hu, Schmitt, and Bulling]{yan2023gazemodiff}
H.~Yan, Z.~Hu, S.~Schmitt, and A.~Bulling.
\newblock Gazemodiff: Gaze-guided diffusion model for stochastic human motion prediction.
\newblock \emph{arXiv preprint arXiv:2312.12090}, 2023.

\bibitem[Chi et~al.(2025)Chi, Xu, Feng, Cousineau, Du, Burchfiel, Tedrake, and Song]{chi2023diffusion}
C.~Chi, Z.~Xu, S.~Feng, E.~Cousineau, Y.~Du, B.~Burchfiel, R.~Tedrake, and S.~Song.
\newblock Diffusion policy: Visuomotor policy learning via action diffusion.
\newblock \emph{The International Journal of Robotics Research}, 44\penalty0 (10-11):\penalty0 1684--1704, 2025.
\newblock \doi{10.1177/02783649241273668}.

\bibitem[Tevet et~al.(2022)Tevet, Raab, Gordon, Shafir, Cohen-Or, and Bermano]{tevet2022human}
G.~Tevet, S.~Raab, B.~Gordon, Y.~Shafir, D.~Cohen-Or, and A.~H. Bermano.
\newblock Human motion diffusion model.
\newblock \emph{arXiv preprint arXiv:2209.14916}, 2022.

\bibitem[{Gemma Team}(2025)]{gemma3_2025}
{Gemma Team}.
\newblock Gemma 3 technical report.
\newblock \emph{arXiv preprint arXiv:2503.19786}, 2025.

\bibitem[Zhao et~al.(2023)Zhao, Ding, An, Du, Yu, Li, Tang, and Wang]{zhao2023fastsam}
X.~Zhao, W.~Ding, Y.~An, Y.~Du, T.~Yu, M.~Li, M.~Tang, and J.~Wang.
\newblock Fast segment anything.
\newblock \emph{arXiv preprint arXiv:2306.12156}, 2023.

\bibitem[Kingma and Ba(2015)]{kingma2015adam}
D.~P. Kingma and J.~Ba.
\newblock Adam: A method for stochastic optimization.
\newblock In \emph{International Conference on Learning Representations}, 2015.

\bibitem[Hochreiter and Schmidhuber(1997)]{hochreiter1997lstm}
S.~Hochreiter and J.~Schmidhuber.
\newblock Long short-term memory.
\newblock \emph{Neural Computation}, 9\penalty0 (8):\penalty0 1735--1780, 1997.

\bibitem[He et~al.(2016)He, Zhang, Ren, and Sun]{he2016resnet}
K.~He, X.~Zhang, S.~Ren, and J.~Sun.
\newblock Deep residual learning for image recognition.
\newblock In \emph{Proceedings of the IEEE Conference on Computer Vision and Pattern Recognition}, pages 770--778, 2016.

\bibitem[Lipman et~al.(2023)Lipman, Chen, Ben-Hamu, Nickel, and Le]{lipman2023flow}
Y.~Lipman, R.~T.~Q. Chen, H.~Ben-Hamu, M.~Nickel, and M.~Le.
\newblock Flow matching for generative modeling.
\newblock In \emph{International Conference on Learning Representations}, 2023.

\bibitem[Varghese and M.(2024)]{10533619}
R.~Varghese and S.~M.
\newblock Yolov8: A novel object detection algorithm with enhanced performance and robustness.
\newblock In \emph{2024 International Conference on Advances in Data Engineering and Intelligent Computing Systems (ADICS)}, pages 1--6, 2024.
\newblock \doi{10.1109/ADICS58448.2024.10533619}.

\bibitem[Lugaresi et~al.(2019)Lugaresi, Tang, Nash, McClanahan, Uboweja, Hays, Zhang, Chang, Yong, Lee, et~al.]{lugaresi2019mediapipe}
C.~Lugaresi, J.~Tang, H.~Nash, C.~McClanahan, E.~Uboweja, M.~Hays, F.~Zhang, C.-L. Chang, M.~G. Yong, J.~Lee, et~al.
\newblock Mediapipe: A framework for building perception pipelines.
\newblock \emph{arXiv preprint arXiv:1906.08172}, 2019.

\bibitem[Freedman(2008)]{freedman2008coordination}
E.~G. Freedman.
\newblock Coordination of the eyes and head during visual orienting.
\newblock \emph{Experimental brain research}, 190\penalty0 (4):\penalty0 369--387, 2008.

\bibitem[Land and Hayhoe(2001)]{land2001ways}
M.~F. Land and M.~Hayhoe.
\newblock In what ways do eye movements contribute to everyday activities?
\newblock \emph{Vision research}, 41\penalty0 (25-26):\penalty0 3559--3565, 2001.

\bibitem[Land et~al.(1999)Land, Mennie, and Rusted]{land1999roles}
M.~Land, N.~Mennie, and J.~Rusted.
\newblock The roles of vision and eye movements in the control of activities of daily living.
\newblock \emph{Perception}, 28\penalty0 (11):\penalty0 1311--1328, 1999.

\end{thebibliography}

\clearpage

\appendix
\section{Appendix and Supplementary Materials}
\paragraph{Code and Data Availability.}
The RGT implementation, markerless collection/preprocessing pipeline, and
MuJoCo simulation documentation are available in the accompanying repository:
\url{https://anonymous.4open.science/r/eye_robot_pub-CF64/README.md}.
The complete participant trajectory dataset is not currently public.

\subsection{Single-step Sliding-Window Evaluation}
\label{app:approach1}

To disentangle one-step prediction quality from long-horizon rollout stability,
we additionally evaluate every learned baseline under a single-step
sliding-window protocol. At each frame $j$ of the held-out portion of a
sequence, the model receives the real history $[j-60, j)$ and the
ground-truth target eyeball rotation for the next chunk, and predicts that
chunk; the window then advances by one frame. Because the history is always
ground truth, there is no error compounding---this protocol measures purely
per-step prediction quality.

Table~\ref{tab:baselines_approach1} reports the resulting joint-angle MSE
($\mathrm{rad}^2$). Under this regime Flow Matching is the strongest model on
five of six behaviors and overall ($1.2\times10^{-5}$, roughly $7\times$ better
than the next-best learned model). This is the inverse of the closed-loop
ranking in Table~\ref{tab:overall_joint_mse}, where Flow Matching is the
weakest learned baseline---a direct demonstration that single-step accuracy
does not transfer to autoregressive deployment, motivating our use of
long-horizon closed-loop rollout as the primary evaluation in the main paper.

\begin{table}[h]
\centering
\caption{Single-step sliding-window evaluation. Per-BID and overall
joint-angle MSE ($\mathrm{rad}^2$). The model always receives ground-truth
history; there is no error compounding. Best per column in bold.}
\scriptsize
\setlength{\tabcolsep}{12pt}
\begin{tabular}{lccccccc}
\toprule
\textbf{Method} & \textbf{BID 1} & \textbf{BID 2} & \textbf{BID 3} & \textbf{BID 4} & \textbf{BID 5} & \textbf{BID 6} & \textbf{Overall} \\
\midrule
RNN                     & 8.6e-5          & 1.3e-4          & 1.2e-4          & 1.9e-4          & 3.2e-4          & 9.9e-4          & 2.6e-4 \\
VAE-CNN                 & 1.2e-5          & 1.5e-5          & 3.7e-5          & \textbf{2.0e-5} & 5.2e-5          & 4.8e-4          & 8.7e-5 \\
Flow Matching (16-step) & \textbf{4.2e-6} & \textbf{1.0e-5} & \textbf{1.4e-5} & 2.6e-5          & \textbf{1.9e-5} & \textbf{1.9e-5} & \textbf{1.2e-5} \\
RGT-Base               & 1.6e-5          & 2.5e-5          & 3.8e-5          & 3.9e-5          & 9.5e-5          & 3.9e-4          & 8.5e-5 \\
\midrule
RGT-Large                  & 1.4e-5          & 2.1e-5          & 4.1e-5          & 2.9e-5          & 8.9e-5          & 6.0e-4          & 1.1e-4 \\
RGT-Small               & 1.4e-5          & 2.3e-5          & 4.3e-5          & 3.2e-5          & 8.5e-5          & 6.5e-4          & 1.2e-4 \\
\bottomrule
\end{tabular}
\label{tab:baselines_approach1}
\end{table}

\subsection{Hardware and Control} \label{sec:hardware_control}
\paragraph{Hardware.} We evaluate our method on a humanoid robotic head~\cite{hu2026lipmotions} with 11 degrees of freedom (DOF), comprising 3 DOFs for the neck (yaw, pitch, and roll), 4 DOFs for the left and right eyes (yaw and pitch per eye), and 4 DOFs for the upper and lower eyelids of both eyes. For each joint, we define motion limits that closely resemble the natural constraints of human anatomy to ensure lifelike articulation. For further details on the robotic platform, we refer the reader to~\cite{hu2026lipmotions}.

\paragraph{Control.} The character is actuated using a set of joint torques $\tau$, computed from the control actions via an actuator model. Given an action input $a_t$, we derive the motor torques using a proportional-derivative (PD) controller:
\begin{equation}
\tau_t^{\text{motor}} = K_P \cdot (a_t - q_t) - K_D \cdot \dot{q}_t,
\end{equation}
where $\cdot$ denotes element-wise multiplication, $q_t$ is the current joint position, and $\dot{q}_t$ is the joint velocity. The proportional and derivative gain terms $K_P$ and $K_D$ are set to predefined values based on the actuator dynamics. In practice, head and eye movements tend to be rapid, so we use relatively high values for $K_P$ to ensure fast and accurate tracking.

\subsection{VLM and Visual Perception Module} \label{sec:Visual Perception Module}

\paragraph{Vision Language Model (VLM). }
The deliberative stage is a Vision-Language Model (\texttt{Gemma~3}) that jointly
decides \emph{where} the robot should attend and \emph{what}, if anything, it
should say. Crucially, it is used \emph{off-the-shelf}: the model is never
fine-tuned, and all task-specific behavior is induced purely through a fixed
prompt template, so the same framework drops in any capable VLM without retraining.
This makes the backend interchangeable---we support both a hosted \textbf{API}
deployment, which minimizes on-board compute and latency, and a fully
\textbf{local} deployment for network-independent, on-device operation---behind an
identical prompt and output contract. Rather than relying on free-form generation,
we cast target selection as a constrained structured-output task. The template
frames the model as a chatting face robot and presents it with two parallel input
streams at every call: a \textbf{VISUAL} stream---the egocentric image $I_t$ from
the eye-embedded cameras---and an \textbf{AUDIO} stream---a possibly-empty
transcript $\ell$ produced by the streaming speech-recognition front end. The
system instruction then specifies the task and the exact output contract:

\begin{quote}\small\itshape
You are a chatting face robot. You receive two parallel streams: (1) VISUAL, an
image; (2) AUDIO, a transcript (possibly empty). Choose the single object or
person in the scene that deserves the most attention given both streams and
return its bounding box as \texttt{[ymin, xmin, ymax, xmax]} normalized to
$[0,1000]$; if the target is a person, return the box of the person's face. If
AUDIO is non-empty, additionally generate a spoken response ($\le 10$\,s); if
AUDIO is empty, omit the speech field. Output \emph{only} JSON.
\end{quote}

This template enforces three properties that the rest of the pipeline depends on.
First, the output is a single JSON object with a bounding-box field and an
optional \texttt{speech} field, which we obtain reliably by requesting a JSON
response type from the model; the structured referent $r_t$ parsed from this object
is precisely the rough region of interest consumed by the downstream perception
module. Second, the normalized $[0,1000]$ box convention decouples the prompt from
the camera resolution, so the same template transfers across the two eye cameras;
we rescale the returned box to pixels before projecting it into the eyeball frame.
Third, instructing the model to return the \emph{face} box for people---rather than
the full body---anchors the eventual fixation on the eyes and preserves the realism
of mutual gaze. Because such reasoning is comparatively expensive, the VLM runs
asynchronously and far below the $60$\,Hz control rate; the reactive perception
module sustains the fixation between successive calls, and the controller always
consumes only the most recent VLM result, discarding stale frames to bound latency.
Unifying target selection and the verbal response within this single templated call
lets one scene understanding drive both gaze and speech, keeping the robot's
attention and dialogue mutually consistent.

\paragraph{Visual Perception Module.}
The robot carries two eye-embedded cameras with different focal lengths: the
left-eye camera is optimized for close-range perception and the right-eye camera
for long-range focus, reliably resolving objects at roughly one meter. The
reactive perception module of Sec.~\ref{sec:vlm_targeting} operates on these
streams, refining the VLM's structured referent $r_t$ into a frame-rate fixation
$u_t \in \mathbb{R}^{2}$ and maintaining it between successive VLM calls.

We use FastSAM~\cite{zhao2023fastsam}, the real-time variant of the Segment
Anything Model, as this module; the choice follows directly from our two-stage
design. Because the upstream VLM selects targets by language and context, the set
of admissible gaze targets is effectively open---any person, object, or region
the VLM can designate---so the downstream localizer must not re-impose a fixed
taxonomy. A class-agnostic, promptable segmenter preserves this open-endedness:
the rough region of interest in $r_t$ is passed directly as a spatial prompt, and
FastSAM returns a pixel-accurate mask of the designated target, whose centroid we
take as $u_t$ and project into the eyeball frame.

In contrast, the detector paradigms commonly used for robot perception fit this
role poorly. Closed-set detectors such as YOLO~\cite{10533619} are confined to a
fixed label set and would bottleneck the VLM's open-vocabulary targeting; their
bounding-box centers are also only coarse aim points, falling on the torso rather
than the face for a standing person, or off the object entirely when it is
elongated or partially occluded---both of which undermine the realism of eye
contact. Category-specific landmark models such as
MediaPipe~\cite{lugaresi2019mediapipe} yield precise points, but only for faces,
hands, and body, and cannot localize arbitrary referents. FastSAM retains the
open-set, prompt-conditioned behavior of SAM while running far faster than its
ViT-based predecessor, placing class-agnostic segmentation comfortably within the
frame-rate budget required to sustain a fixation between VLM calls and thereby
preserve the perceived liveness of the robot's gaze.

\subsection{Dataset Collection, Description and Representation} \label{sec:dataset_description}

Our collection pipeline is markerless and uses a commodity webcam. Participants
were recruited for a non-clinical feasibility study. We first record raw RGB video of a seated
participant and then process every frame offline with
the MediaPipe Face Landmarker~\cite{lugaresi2019mediapipe}, run in video mode
with face-blendshape and facial-transformation-matrix outputs enabled. The 11-DoF
target representation is read out directly from these outputs. Head orientation
(yaw, pitch, roll) is recovered as Euler angles from the $4\times4$ facial
transformation matrix. Per-eye gaze is obtained from the directional eye
blendshapes---horizontal gaze from the difference of the \texttt{eyeLookIn} and
\texttt{eyeLookOut} scores and vertical gaze from \texttt{eyeLookUp} minus
\texttt{eyeLookDown}, computed independently for each eye---and linearly scaled to
anatomically plausible ranges ($\pm30^\circ$ yaw, $\pm20^\circ$ pitch). Upper-eyelid
aperture is mapped from the \texttt{eyeBlink} blendshape and lower-eyelid motion
from \texttt{eyeSquint}, each calibrated to the robot's eyelid travel. Each frame
is timestamped, so the subsequent resampling stage (Sec.~\ref{sec:Preprocessing})
can interpolate onto a uniform 60\,FPS grid regardless of capture rate. Because the
entire pipeline relies only on pretrained MediaPipe models, new subjects or
behaviors can be added without any specialized motion-capture hardware. Details of the motion extraction process are provided in the later section \ref{sec:Preprocessing}.

The data used here are organized as S1, six structured behavior sequences from
Subject~1, and S2, five participant-specific adaptation sequences from
Subjects~2--6. S1 is used for pretraining and the primary
benchmark; S2 is used for participant-wise adaptation. The recordings were
designed to elicit tracking, scanning, surveying, and individual motion patterns.
The counts in the S1 behavior summary below are processed 60-Hz command-grid
rows rather than raw camera frames.

The six structured motion behaviors are summarized in Table~\ref{tab:motion_summary}. Each behavior is designed to reflect a distinct style of gaze deployment, varying in head amplitude, eye-head contribution, and overall dynamics.
\begin{table}[h!]
\centering
\caption{Summary of S1 human motion data; counts are processed 60-Hz rows.}
\begin{tabularx}{\textwidth}{|c|X|c|}
\hline
\textbf{Behavior (Subject)} & \textbf{Description} & \textbf{Rows} \\
\hline
Behavior 1 (Subject 1) & This sequence mimics natural human \textbf{eye tracking} behavior. The head oscillates within $\pm$10$^\circ$, while the eyes move in coordination, turning left and right with an anticipatory expression. & 31487 \\
\hline
Behavior 2 (Subject 1) & Similar to Behavior 1, this sequence replicates \textbf{eye tracking} behavior but over a shorter duration. The head and eye motions follow the same pattern. & 26739 \\
\hline
Behavior 3 (Subject 1) & This sequence captures a \textbf{Head-Down Scan} gesture. The motion begins by lowering the head approximately 30$^\circ$, followed by a slow left-right head rotation of about 15$^\circ$, expressing contemplation. & 8358 \\
\hline
Behavior 4 (Subject 1) & This sequence represents a \textbf{Head-Up Survey} gesture. The head raises approximately 45$^\circ$ and then moves left and right, coordinated with eye movements, expressing alertness. & 6750 \\
\hline
Behavior 5 (Subject 1) & This sequence features a \textbf{Head-Mid Survey}, using head oscillations within $\pm$10$^\circ$ along with coordinated eye movements. & 10014 \\
\hline
Behavior 6 (Subject 1) & This sequence follows an \textbf{All Survey} pattern, combining head oscillations within $\pm$10$^\circ$ and coordinated eye movements to express anticipation. & 14310 \\
\hline
\end{tabularx}
\label{tab:motion_summary}
\end{table}
Each motion sequence is represented as a time series of 11 joint angles, normalized using the global mean and standard deviation computed over the full training set. This normalization ensures consistent scaling across different joints and motion types, facilitating stable learning and evaluation. 

The complete dataset is not currently public; the available repository and
current release scope are described in the Code and Data Availability paragraph.

\subsection{Data Preprocessing Details}
\label{sec:Preprocessing}

Let the raw motion sequence be denoted as \( D = \{ (t_i, x_i) \}_{i=1}^{T_{\text{raw}}} \), where each frame \( x_i \in \mathbb{R}^{11} \) contains head orientation (yaw, pitch, roll), binocular eye gaze (yaw and pitch for each eye), and upper and lower eyelid motions.

\paragraph{Temporal Resampling.}  
To ensure uniform temporal spacing, we perform linear interpolation to resample \(D\) at a constant frame rate of 60 frames per second (FPS), producing a sequence \( \hat{D} = \{ (t'_j, \hat{x}_j) \}_{j=1}^{T} \), where \( t'_j = t'_1 + (j-1)\Delta t \) and \( \Delta t = 1/60 \) seconds.

\paragraph{Eye Gaze Denoising.}  
For selected eye gaze channels \( \mathcal{G} \subseteq \{4,5,6,7\} \), spike artifacts are removed via peak detection. Peaks are identified where the signal magnitude exceeds a dynamic threshold, and values within a window around each detected peak are masked and reconstructed by linear interpolation.  
Subsequently, a Gaussian smoothing filter is applied to the cleaned eye gaze channels:
\[
\tilde{x}_j^{(g)} = (G_{\sigma} * \hat{x}^{(g)})_j, \quad g \in \mathcal{G}
\]
where \(G_{\sigma}\) denotes a 1D Gaussian kernel with standard deviation \(\sigma\).

\paragraph{Eyelid Denoising.}
For the eyelid channels \( \mathcal{E} \subseteq \{8,9,10,11\} \), the dip
produced by each detected blink is first flattened by replacing the blink
samples with linear interpolation across the blink (each upper eyelid is
flattened using its own blink detection, and each lower eyelid reuses the
corresponding upper eyelid as the blink source, since the lower lids move
little during a blink). The blink shape is reintroduced later at synthesis
time (Sec.~\ref{sec:blink}). The blink-flattened channels are then lightly
denoised with a Gaussian smoothing filter,
\[
\tilde{x}_j^{(e)} = (G_{\sigma_e} * \hat{x}^{(e)})_j, \qquad e \in \mathcal{E},
\]
with \(\sigma_e = 2\) frames, so that non-blink samples receive mild smoothing
while the flattened ramps remain flat. This yields the blink-free eyelid
signal on which the reconstruction objective is trained.

\paragraph{Blink Label Extraction.}  
Blink labels \( P_t \) are extracted from the upper eyelid motion channel. Blinks are detected as local minima surpassing a negative threshold.  
Binary blink labels are assigned as:
\[
P_t = 
\begin{cases}
1, & \text{if } t \in \text{blink window around detected peak},\\
0, & \text{otherwise}.
\end{cases}
\]
Soft blink labels are generated by convolving the detected blink events with a Gaussian kernel to model temporal uncertainty and smoothness:
\[
\tilde{P}_t = \sum_{k \in \mathcal{P}} \exp\left( -\frac{(t-k)^2}{2\sigma^2} \right)
\]
where \(\mathcal{P}\) is the set of detected blink peak indices.

\subsection{Blink Template Extraction and Synthesis}
\label{sec:blink}

Blink events are sparse relative to non-blink frames, which motivates a
separate timing and playback path in addition to the reconstruction loss.
We therefore adopt a two--step strategy:  
(1)~offline extraction of blink templates from the markerless recording set, and
(2)~online synthesis that injects these templates at run-time according to the blink--probability head of our model.

\paragraph{Template extraction.}
A blink is deemed strong when the signal crosses a fixed threshold of $-0.8$ (normalised units) and forms a local minimum separated from neighbouring candidates by at least 30 frames ($\simeq\!0.5$ s).Around every detected centre we crop a $20$-frame window ($\pm\,166$ ms) comprising all four eyelid channels, yielding a tensor $\mathbf{T}\in\mathbb{R}^{N\times20\times4}$ of blink templates (``open\,$\rightarrow$\,closed\,$\rightarrow$\,open''). Across six subjects we obtain $N=1\,{,}073$ valid blinks.  
This corpus serves two purposes: (i) supervision for the auxiliary blink head, and (ii) a library for synthesis.

\paragraph{Blink-probability supervision.}
Binary blink labels are first assigned at peak frames and then convolved with a Gaussian kernel ($\sigma\!=\!3$\,frames) to produce smooth targets \(P_t^{\star}\).  
During training we optimise a weighted binary cross-entropy
$\lambda_{\text{BCE}}\operatorname{BCE}(P_t,P_t^{\star})$ where positive blink frames receive unit weight and non-blink frames are down-weighted, alleviating severe class imbalance.

\paragraph{Probabilistic blink synthesis.}
At inference time RGT outputs a per-frame blink probability
\(\hat{P}_t\!\in\![0,1]\). A blink is triggered when \(\hat{P}_t\) exceeds
$0.72$, subject to a refractory period of $100$ frames ($\sim$1.7\,s) that
prevents unnaturally rapid blinking. Once triggered, we draw a random
template \(\mathbf{T}_k\) from the library and play it back frame by frame
over its $20$-frame window, substituting the four template eyelid channels
directly for the network's eyelid outputs,
\[
\mathbf{E}_{\tau+j} \leftarrow \mathbf{T}_{k,j}, \qquad j = 0,\dots,19,
\]
while the head and gaze channels continue to follow the generated
trajectory. When the window completes, the eyelids hand control back to the
network outputs and a fresh template is sampled for the next blink.

\subsection{Data Analysis}

\paragraph{Observation 1: Eye--Head Co-movement}

\begin{figure}[H]
    \centering

    \begin{subfigure}[b]{\textwidth}
        \centering
        \includegraphics[width=0.8\textwidth]{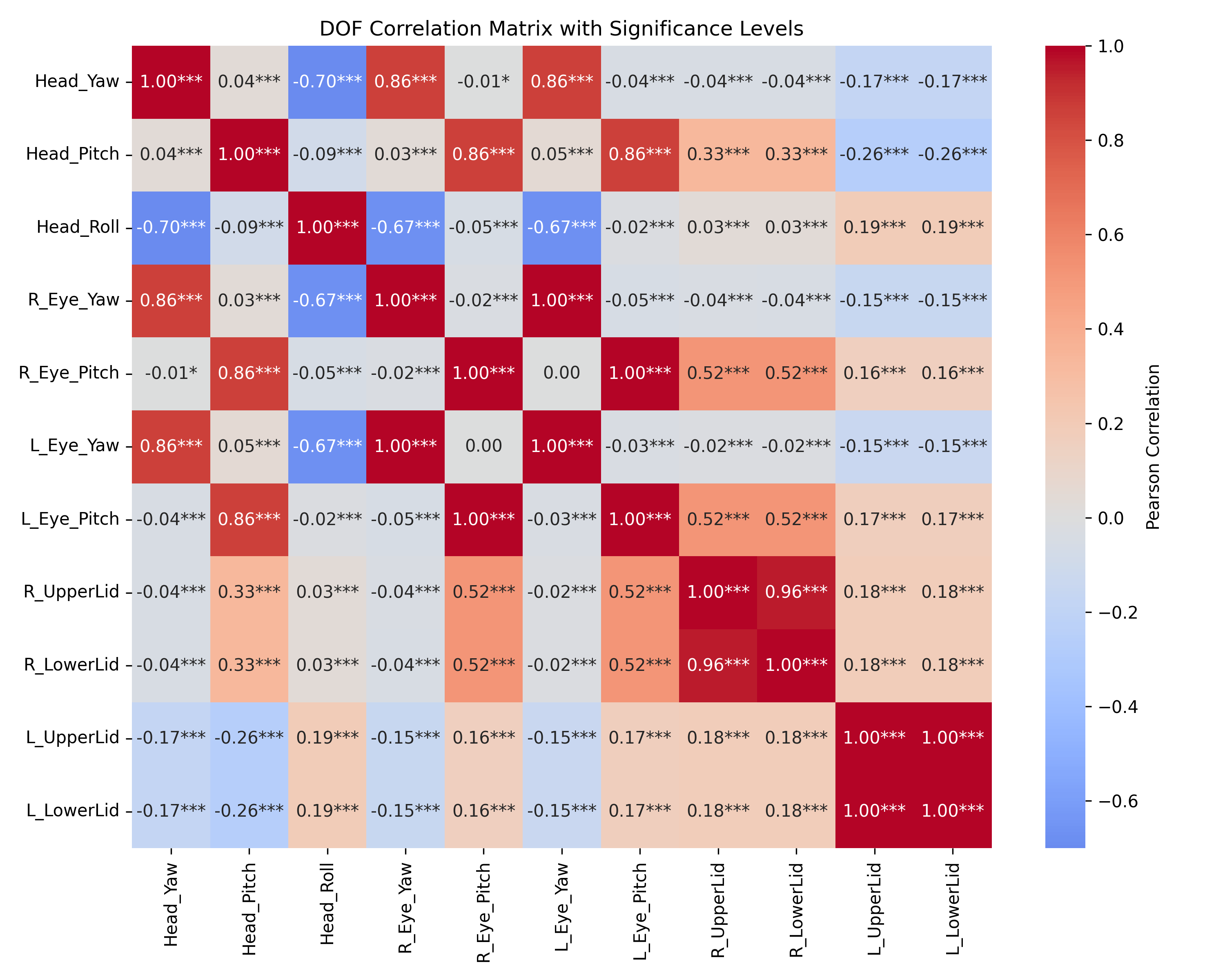}
        \caption{
            Pearson correlation matrix across head yaw/pitch, eye yaw/pitch, and eyelid motions, computed over the full motion dataset.
        }
        \label{fig:full_dof_corr}
    \end{subfigure}

    \vspace{1em}

    \begin{subfigure}[b]{1.0\textwidth}
        \centering
        \includegraphics[width=0.8\textwidth]{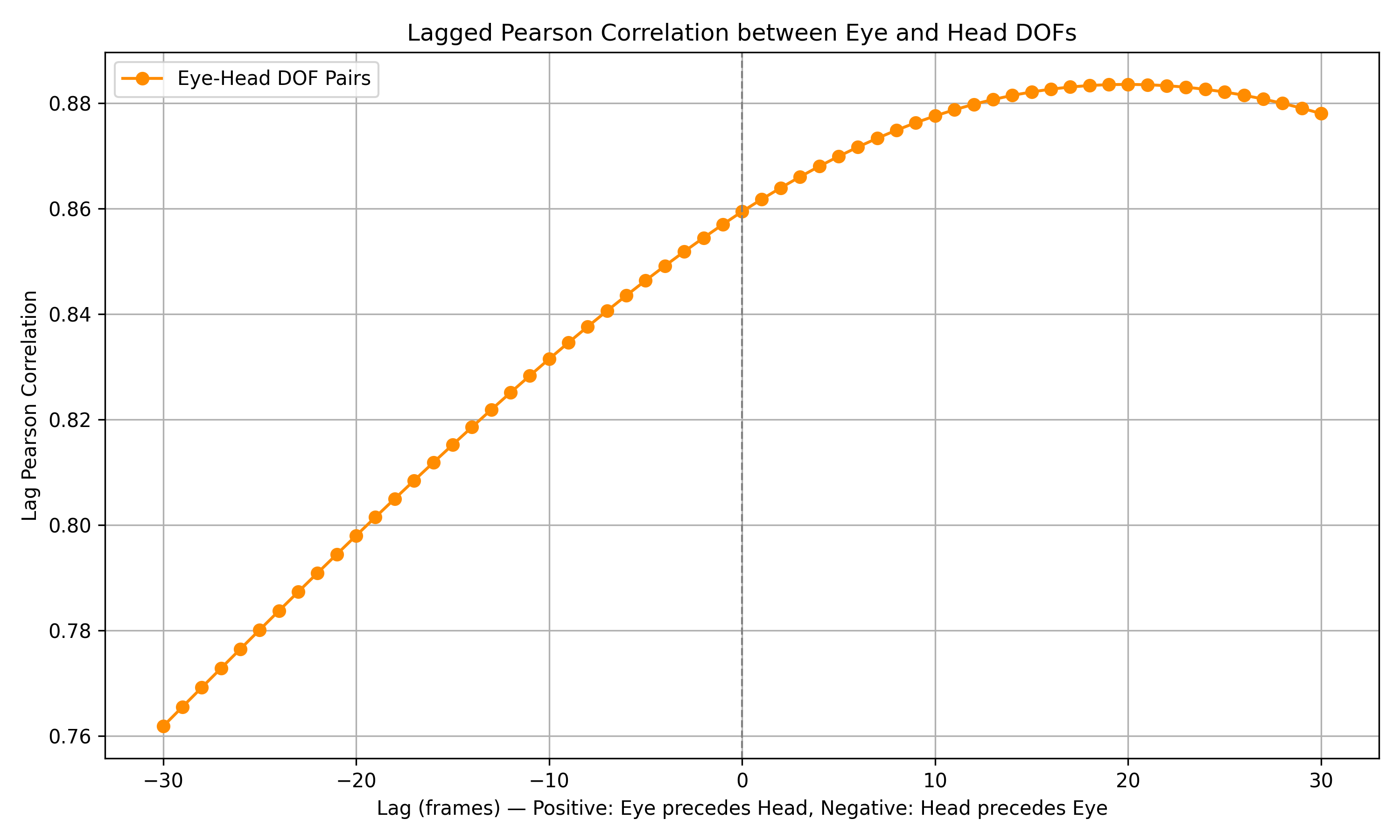}
        \caption{
            Lagged Pearson correlation between key eye and head DOFs (head yaw/pitch vs. eye yaw/pitch). Positive lag values indicate that the eye leads the head, while negative values indicate the head leads the eye.
        }
        \label{fig:Corr_Lag}
    \end{subfigure}

    \caption{
        (Top) Pearson correlation matrix across all head, eye, and eyelid degrees of freedom (DOFs), revealing strong coupling between head yaw and eye yaw, as well as head pitch and eye pitch.
        (Bottom) Lagged Pearson correlation peaks at a positive lag of approximately 20 frames (333 ms), consistent with eye motion preceding head motion in these recordings. Per-frame significance labels shown within the supplied plots are exploratory because adjacent frames are temporally correlated.
    }
    \label{fig:combined_correlation}
\end{figure}

Prior human motor-control studies report coupling between head and eye movement
during gaze shifts~\cite{freedman2008coordination,land2001ways}. We summarize the
linear associations among processed head, eye, and eyelid trajectories using
Pearson correlation $\rho$. Figure~\ref{fig:full_dof_corr} shows an eye--head
correlation of $\rho=0.86$ for the reported yaw/pitch comparison. These
descriptive correlations characterize the collected kinematics; temporal
autocorrelation means a naive independent-frame significance test would be
inappropriate.

\clearpage

\paragraph{Observation 2: Eye Yaw (Pitch) Leads Head Yaw (Pitch).}
To examine the temporal relationship during gaze shifts, we compute lagged
Pearson correlation across the eye and head yaw/pitch channels. For a lag
$\mathbf{L}^n$, $n$ is the number of 60-Hz frames by which head motion follows
eye motion, and the correlation is denoted $\rho^n$.

Figure~\ref{fig:Corr_Lag} presents these lagged correlations. Positive lag
values indicate earlier eye motion; the reported maximum is at
$\mathbf{L}^{20}$ (333 ms), with $\rho^{20}=0.88$. This pattern is consistent
with earlier eye motion in these sequences and with previous observations of
human gaze coordination~\cite{land1999roles}.

The RGT conditions predicted neck motion on eye commands and recent full-head
history to model this temporal association.

\subsection{Online Generation of the Eyeball-Condition Signal}
\label{sec:eye_cond_online}

During real-robot experiments the 4-D ocular command \(C_t\in\mathbb{R}^{f\times4}\) is produced on-the-fly by an Anchor--Interpolation Trajectory Generator that is both fully deterministic and sub-millisecond on a single CPU core.  
A stream of 3-D gaze targets
\(\mathbf g_1,\mathbf g_2,\dots,\mathbf g_m\in\mathbb{R}^{3}\) arrives from the perception module; each \(\mathbf g_i\) is projected into binocular eye space to yield a pitch--yaw pair for the right and left eye,
\[
\mathbf a_i=\bigl[\,
(\theta^{\mathrm R}_{\text{pitch}},\theta^{\mathrm R}_{\text{yaw}}),
(\theta^{\mathrm L}_{\text{pitch}},\theta^{\mathrm L}_{\text{yaw}})
\bigr]^{\!\top}\in\mathbb{R}^{4}.
\]

\paragraph{Anchor queue and dwell times.}
Every new anchor \(\mathbf a_i\) is appended to a rolling queue together with a dwell time
\(\tau_i\sim\mathcal{U}[\tau_{\min},\tau_{\max}]\)  
with \(\tau_{\min}=0.2\,\mathrm s\) and \(\tau_{\max}=1.0\,\mathrm s\).
At 60\,fps this corresponds to \(n_i=\lceil\tau_i\!\cdot\!60\rceil\) repeated frames.

\paragraph{Linear interpolation between anchors.}
Let \(\mathbf a_{i}\) and \(\mathbf a_{i+1}\) be two consecutive anchors.
We generate \(f=20\) uniformly spaced interpolation samples
\(\mathbf c_{i,k}\;(k=1\ldots f)\) by
\[
\mathbf c_{i,k}
      \;=\;
      \bigl(1-\alpha_k\bigr)\,\mathbf a_{i}
      +\alpha_k\,\mathbf a_{i+1},
      \quad
      \alpha_k=\frac{k}{f}.
\]
This produces a \(t\approx 0.33\,\mathrm s\) transition that is perceptually smooth yet fast enough for conversational gaze.

\paragraph{Velocity cap and mechanical limits.}
Each increment
\(\dot{\mathbf c}_{i,k}=(\mathbf c_{i,k}-\mathbf c_{i,k-1})/0.0167\,\mathrm s\)
is clipped to a hard speed bound
\(\lVert\dot{\mathbf c}_{i,k}\rVert_\infty\le400^{\circ}\!/\mathrm s\).
The resulting angles are further clipped to the mechanical limits of the ocular joints (Appendix~\ref{sec:hardware_control}).

\paragraph{Stochastic target perturbations.}
For optional variation in the procedural eye-condition stream, we splice in
\(S\sim\mathcal{U}\{1,2,3\}\) short target bursts.
For each burst we pick an insertion point along the current trajectory and draw \(K\sim\mathcal{U}\{1,\dots,5\}\) offset targets uniformly from
\([-0.6,+0.6]^{2}\)rad, and render each as a square pulse of 5--10 frames.
These pitch--yaw perturbations inherit the same speed and joint-limit checks.

\paragraph{Outcome.}
The generator outputs a continuous stream of 4-D ocular commands with \(<1\) ms latency, leaving all head and eyelid coordination to the RGT decoder.  
This division of labour keeps the control stack simple and allows RGT to focus entirely on synthesising lifelike full-head motion given a task-directed, but otherwise non-parametric, eye trajectory.

\end{document}